\pdfoutput=1

\documentclass[11pt]{article}

\usepackage[]{acl}

\usepackage{times}
\usepackage{latexsym}

\usepackage[T1]{fontenc}

\usepackage[utf8]{inputenc}

\usepackage{microtype}

\usepackage{inconsolata}

\usepackage{amsmath}
\usepackage{times}
\usepackage{framed}
\usepackage{ragged2e}
\usepackage{soul}
\usepackage{footmisc}
\usepackage{latexsym}
\usepackage{subfigure}
\usepackage{amsthm}
\usepackage{graphicx}
\usepackage{float}
\usepackage{longtable}
\usepackage{booktabs} 
\usepackage{multirow}
\usepackage{multicol}
\usepackage{amssymb}
\usepackage{dashbox}%
\usepackage{multirow}
\usepackage{textcomp}
\usepackage{tcolorbox}
\usepackage{bbding}
\usepackage[ruled, vlined, linesnumbered]{algorithm2e}
\usepackage{mathtools}
\usepackage{adjustbox}
\usepackage[ruled,vlined]{algorithm2e}
\usepackage{color, soul}
\usepackage{array}
\newcolumntype{P}[1]{>{\centering\arraybackslash}p{#1}}
\newcolumntype{M}[1]{>{\centering\arraybackslash}m{#1}}
\usepackage{cleveref}
\usepackage{colortbl}
\crefname{section}{§}{§§}
\Crefname{section}{§}{§§}
\crefname{figure}{Figure}{Figure}
\Crefname{figure}{Figure}{Figure}
\crefname{table}{Table}{Table}
\Crefname{table}{Table}{Table}
\definecolor{forestgreen}{HTML}{228B22}
\newcommand\mycheck{\textcolor{forestgreen}{\Checkmark}\xspace}
\newcommand\myx{\textcolor{red}{\XSolidBrush}\xspace}
\newcommand\ourdata{\textsc{Maven-Arg}\xspace}
\newcommand\mavenere{\textsc{Maven-Ere}\xspace}
\newcommand\maven{\textsc{Maven}\xspace}

\definecolor{'wit'}{HTML}{FBFBFB}
\definecolor{'gry'}{HTML}{EEEEEE}

\definecolor{'deep1'}{HTML}{C5E6F8} 
\definecolor{'shallow1'}{HTML}{E4F3FC} 
\definecolor{'deep2'}{HTML}{E5F5B7} 
\definecolor{'shallow2'}{HTML}{F3FADF} 

\definecolor{'deep3'}{HTML}{FFE5C6} 
\definecolor{'shallow3'}{HTML}{FFF2E3} 
\definecolor{'deep4'}{HTML}{FFD3CF} 
\definecolor{'shallow4'}{HTML}{FFEAE8}
\definecolor{'deep5'}{HTML}{D2D0F3} 
\definecolor{'shallow5'}{HTML}{E8E7F9}

\title{\ourdata: Completing the Puzzle of All-in-One Event Understanding Dataset with Event Argument Annotation}

\author{ Xiaozhi~Wang$^{1}$, Hao~Peng$^{1}$, Yong~Guan$^{1}$, Kaisheng~Zeng$^{1}$, Jianhui~Chen$^{1}$, Lei~Hou$^{1,2}$, \\ \textbf{Xu~Han$^{1}$, Yankai~Lin$^{3}$, Zhiyuan~Liu$^{1,2}$, Ruobing~Xie$^{4}$, Jie~Zhou$^4$, Juanzi Li$^{1,2}$\thanks{\quad Corresponding author.}} \\
 $^1$Department of Computer Science and Technology, BNRist;\\$^2$KIRC, Institute for Artificial Intelligence,\\Tsinghua University, Beijing, China\\
 $^3$Gaoling School of Artificial Intelligence, Renmin University of China, Beijing, China \\
 $^4$WeChat AI, Tencent Inc, China \\
 \texttt{wangxz20@mails.tsinghua.edu.cn}
 }

\begin{document}
\maketitle
\begin{abstract}
Understanding events in texts is a core objective of natural language understanding, which requires detecting event occurrences, extracting event arguments, and analyzing inter-event relationships. However, due to the annotation challenges brought by task complexity, a large-scale dataset covering the full process of event understanding has long been absent. In this paper, we introduce \ourdata, which augments \maven datasets with event argument annotations, making the first all-in-one dataset supporting event detection, event argument extraction (EAE), and event relation extraction. As an EAE benchmark, \ourdata offers three main advantages: (1) a \textbf{comprehensive schema} covering $162$ event types and $612$ argument roles, all with expert-written definitions and examples; (2) a \textbf{large data scale}, containing $98,591$ events and $290,613$ arguments obtained with laborious human annotation; (3) the \textbf{exhaustive annotation} supporting all task variants of EAE, which annotates both entity and non-entity event arguments in document level. Experiments indicate that \ourdata is quite challenging for both fine-tuned EAE models and proprietary large language models (LLMs). Furthermore, to demonstrate the benefits of an all-in-one dataset, we preliminarily explore a potential application, future event prediction, with LLMs. \ourdata and codes can be obtained from \url{https://github.com/THU-KEG/MAVEN-Argument}.
\end{abstract}

\section{Introduction}
\begin{figure}[t!]
    \centering
    \includegraphics[width=0.99\linewidth]{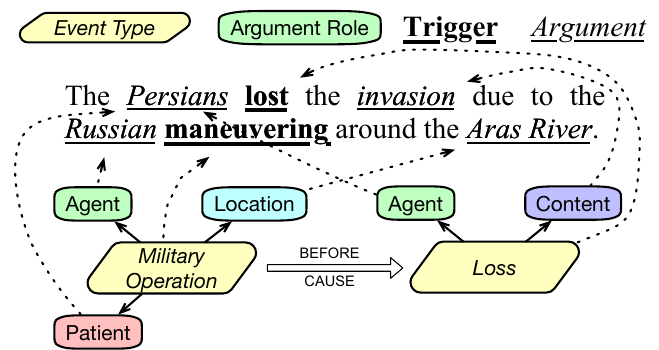}
    \caption{Illustration for the overall event understanding, consisting of event detection, event argument extraction, and event relation extraction tasks.}
    \label{fig:tasks_and_datasets}
\end{figure}

Conveying information about events is a core function of human languages~\citep{levelt1993speaking,pinker2013learnability,miller2013language}, which highlights \textit{event understanding} as a major objective for natural language understanding and a foundation for various downstream applications~\citep{ding2015deep,DBLP:conf/ijcai/LiDL18,goldfarb-tarrant-etal-2019-plan,huang-etal-2019-cosmos,wang2021incorporating}. As illustrated in \cref{fig:tasks_and_datasets}, event understanding is typically organized as three information extraction tasks~\citep{ma-etal-2021-eventplus,peng2023omnievent}: event detection (ED), which detects event occurrences by identifying event triggers and classifying event types; event argument extraction (EAE), which extracts event arguments and classifies their argument roles; event relation extraction (ERE), which analyzes the coreference, temporal, causal, and hierarchical relationships among events.

Despite the importance of event understanding, a large-scale dataset covering all the event understanding tasks has long been absent. Established sentence-level event extraction (ED and EAE) datasets like ACE 2005~\citep{walker2006ace} and TAC KBP~\citep{ellis2015overview,ellis2016overview,getman2017overview} do not involve event relation types besides the basic coreferences. RAMS~\citep{rams} and WikiEvents~\citep{wikievents} extend EAE to the document level but do not involve event relations. ERE datasets are mostly developed independently for coreference~\citep{cybulska-vossen-2014-using}, temporal~\citep{chambers-etal-2014-dense,ning2018matres}, causal~\citep{mirza-etal-2014-annotating,mostafazadeh-etal-2016-caters,caselli-vossen-2017-event}, and subevent~\citep{hovy-etal-2013-events,glavavs2014event} relationships and do not cover event arguments. Given annotation challenges from task complexity, these datasets often cover only thousands of events. Due to the inconsistent event schemata and data, these datasets cannot be unified. This status quo hinders the development of end-to-end event understanding methods and limits the potential for event-based downstream applications. 

\maven~\citep{wang-etal-2020-maven} is the largest human-annotated ED dataset, with a high-coverage event schema for general-domain events. Based on it, \citet{wang2022maven} further annotates the first unified ERE dataset \mavenere, which covers all four types of event relationships and has a massive scale with more than one million event relations. Building on the sustained efforts of these works over years, we complete the puzzle of an all-in-one event understanding dataset in this work. We construct \ourdata, which provides exhaustive event argument annotations based on \maven.

Beyond finishing an all-in-one event understanding dataset, three main advantages of \ourdata make it a valuable EAE benchmark. (1) \textbf{Comprehensive Event Schema.} The original \maven schema only defines event types but without argument roles. We engage experts to enhance MAVEN schema with argument roles and to write detailed definitions for them, which help annotators and can also serve as task instructions for prompting large language models. The resulting event schema contains $162$ event types, $612$ argument roles, and $14,655$ words of definitions, which well cover general-domain events. (2) \textbf{Large Data Scale.} \ourdata comprises $107,507$ event mentions, $290,613$ event arguments, and $129,126$ entity mentions, all of which are human annotated. To our knowledge, this makes it the largest EAE dataset currently available. (3) \textbf{Exhaustive Annotation.} The development of EAE has seen many variations in task settings, including annotating only the topic event~\citep{rams,tong-etal-2022-docee} of a document or all fine-grained events~\citep{walker2006ace}, annotating event arguments at the sentence level~\citep{walker2006ace} or document level~\citep{rams,wikievents}, and limiting event arguments to entities~\citep{walker2006ace,wikievents} or including non-entity arguments~\citep{grishman-sundheim-1996-message,parekh-etal-2023-geneva}. \ourdata adopts the most exhaustive annotation. We annotate event arguments for all fine-grained events at the document level, covering both entity and non-entity arguments. This enhances the dataset's utility for benchmarking and developing a wide range of EAE methods.

In the experiments, we reproduce several recent state-of-the-art EAE models as baselines and also evaluate large language models with in-context learning. Experimental results show that they can only achieve at most $40$\% F1 scores, which is far from promising. It indicates that \ourdata is quite challenging and more research efforts are needed to develop practical EAE methods. Furthermore, to demonstrate the advantage of an all-in-one event understanding dataset for enabling sophisticated event-based applications, we conduct a preliminary exploration of \textit{future event prediction}. We sample causally related event chains from \ourdata and prompt LLMs to predict future events, including their types and arguments. Experiments show that while most of the predictions are reasonable, they seldom align with the actual future. We encourage future work to further explore this application and hope \ourdata can help improve EAE and develop diverse event-based applications.

\section{Dataset Construction}

In this section, we introduce the dataset construction methodology of \ourdata, which involves designing event schema (\cref{sec:schema}), annotating entities (\cref{sec:entity_annotation}) and event arguments (\cref{sec:arg_annotation}).
\subsection{Event Schema Creation}
\label{sec:schema}
The event schema of \maven~\citep{wang-etal-2020-maven} covers a broad range of general-domain events and has a well-defined hierarchical structure. To enable event argument annotation based on \maven, one author and two engaged linguistic experts devoted three years to manually designing argument roles for \maven schema. Each argument role is accompanied by informative text definitions that are easy to understand, and each event type is provided with detailed annotation examples. An example in shown in \cref{app:annotation_struction}. This not only helps annotators understand their tasks but also can prompt LLMs to perform EAE via in-context learning. To ensure quality, the argument role design for each event type is reviewed by at least one expert.

Our event schema creation involves the following steps: (1) Initially, to reduce annotation difficulty, we invite ten ordinary annotators, who are without dedicated study on event semantics, to review the event type schema and a portion of the data. Based on their feedback, we deleted $6$ event types that are similar to others and renamed $4$ event types for clarity. (2) The basic schema is constructed from a simplification and modification of FrameNet~\citep{baker1998berkeley}. The \textit{frame elements} in FrameNet are widely considered akin to argument roles~\citep{aguilar-etal-2014-comparison,parekh-etal-2023-geneva}, but they are often too complex for ordinary annotators to comprehend since FrameNet is primarily constructed for linguistic experts~\citep{aguilar-etal-2014-comparison}. Therefore, for each event type, we manually select frame elements related to describing events and suitable for annotation as \ourdata argument roles from their FrameNet equivalents, and we rewrite the definitions and examples. (3) Extending argument roles based on the 5W1H analysis (What, Where, When, Why, Who, How) for describing events~\citep{DBLP:conf/siu/KaramanYO17,DBLP:conf/recsys/HamborgBG19}. Temporal and causal relations from event relation extraction describe When and Why, while the event type describes What. We primarily refer to Who (participants), Where (locations), and How (manners, instruments, etc.) to design argument roles. (4) Considering the hierarchical structure. When designing subordinate types, we inherit and refine the argument roles of their superordinate types. (5) Sampling data to check if any event argument is missing.

\paragraph{Schema Statistics}
\begin{table}[t!]
\centering
    \begin{adjustbox}{max width=0.9\linewidth}
{
\begin{tabular}{lrr}
\toprule
\textbf{Dataset}  & \multicolumn{1}{c}{\textbf{\#Event Type}} & \multicolumn{1}{c}{\textbf{\#Argument Role}} \\
\midrule
ACE 2005   & $33$                                     & $36$                                        \\
DocEE      & $59$                                     & $356$                                       \\
WikiEvents & $50$                                     & $59$                                        \\
RAMS       & $139$                                    & $65$                                        \\
MEE        & $16$                                     & $23$                                        \\
GENEVA     & $115$                                    & $220$                                       \\
\midrule
\ourdata   & $162$                                    & $612$                                      \\
\bottomrule
\end{tabular}
}
\end{adjustbox}
\caption{Event schema statistics of \ourdata compared with other datasets.}
\label{tab:schema_stat}
\end{table}

After the schema design, the final \ourdata schema contains $162$ event types, $612$ unique argument roles, and $14,655$ words of definitions. Taking inspiration from semantic role labeling~\citep{fillmore1976frame,banarescu-etal-2013-abstract}, we tend to let the argument roles sharing the same semantic role use the same name but distinguish them with different textual definitions. For instance, we do not use \texttt{Killer} for the \texttt{Killing} event type and use \texttt{Attacker} for the \texttt{Attack} event type. Instead, we use \texttt{Agent} to denote them both but write different definitions for them. This is to encourage the knowledge transfer between EAE for different event types. Therefore, $612$ is the number of argument roles with unique definitions, and there are $143$ unique names for all the argument roles. \cref{tab:schema_stat} compares the event schema size of \ourdata with existing EAE datasets, including ACE 2005~\citep{walker2006ace}, DocEE~\citep{tong-etal-2022-docee}, WikiEvents~\citep{wikievents}, RAMS~\citep{rams}, MEE~\citep{pouran-ben-veyseh-etal-2022-mee}, and GENEVA\footnote{GENEVA has a larger ontology without data. Here we compare with its schema actually used in the dataset.}~\citep{parekh-etal-2023-geneva}. We can observe that \ourdata has the largest event schema, which more comprehensively covers the broad range of diverse events and will help develop more generalizable methods.

\subsection{Entity Annotation}
\label{sec:entity_annotation}

The mainstream task setting for EAE~\citep{walker2006ace,wikievents} confines event arguments to entities, which reduces the task's complexity to some extent and provides more definite and standardized extraction results. Hence, before annotating event arguments, we annotate entities for the $4,480$ \maven documents. We follow the task definition and guidelines of a recent named entity recognition benchmark Few-NERD~\citep{ding-etal-2021-nerd}, but we only annotate coarse-grained entity types, including \texttt{Person}, \texttt{Organization}, \texttt{Location}, \texttt{Building}, \texttt{Product}, \texttt{Art}, and \texttt{MISC}. To deliver more unambiguous EAE results and reduce the argument annotation difficulty, we also annotate entity coreference, which means judging whether multiple entity mentions refer to the same entity.

During entity annotation, we engage $47$ annotators, including $8$ senior annotators selected during the annotation training. Each document is annotated by three independent annotators and further checked by one senior annotator. The final annotation results are aggregated via majority voting. If the senior annotator judged the accuracy of a document's annotation to be below $90$\%, the document will be returned to the three first-stage annotators for re-annotation. To check data quality, we calculate Fleiss' kappa~\citep{fleiss1971measuring} to measure the inter-annotator agreements. The result for entity recognition is $73.2$\%, and for entity coreference is $78.4$\%, both indicating high consistency.

\begin{table*}[t!]
\centering
    \begin{adjustbox}{max width=1\linewidth}
{
\begin{tabular}{lrrrrrrccrc}	
\toprule
\multicolumn{1}{l}{\textbf{Dataset}} & \multicolumn{1}{c}{\textbf{\#Doc.}} & \multicolumn{1}{c}{\textbf{\#Event}} & \multicolumn{1}{c}{\textbf{\#Trigger}} & \multicolumn{1}{c}{\textbf{\#Arg.}} & \multicolumn{1}{c}{\textbf{\#Entity}} & \multicolumn{1}{c}{\textbf{\begin{tabular}[c]{@{}c@{}}\#Entity\\ Mention\end{tabular}}} & \textbf{\begin{tabular}[c]{@{}c@{}}Fine-grained \\ Event\end{tabular}} & \textbf{\begin{tabular}[c]{@{}c@{}}Doc.\\ Level\end{tabular}} & \textbf{\begin{tabular}[c]{@{}c@{}}Entity \\ Arg.\end{tabular}} & \textbf{\begin{tabular}[c]{@{}c@{}}Non-Entity \\ Arg.\end{tabular}} \\
\midrule 
ACE 2005         & $599$                              & $4,090$                             & \multicolumn{1}{r}{$5,349$}   & $9,683$                            & \multicolumn{1}{r}{$45,486$} & $59,430$                                                                               & \mycheck                                                               & \myx                                                          & $9,683$                                                                             & \myx                                                                \\
DocEE            & $27,485$                           & $27,485$                            & -                             & $180,528$                          & -                            & \multicolumn{1}{r}{-}                                                                  & \myx                                                                   & \mycheck                                                      & \multicolumn{1}{c}{\myx}                                                            & \multicolumn{1}{r}{$180,528$}                                       \\
WikiEvents       & $246$                              & $3,951$                             & -                             & $5,536$                            & \multicolumn{1}{r}{$13,937$} & $33,225$                                                                               & \mycheck                                                               & \mycheck                                                      & $5,536$                                                                             & \myx                                                                \\
RAMS             & $3,993$                            & $9,124$                             & -                             & $21,237$                           & -                            & \multicolumn{1}{r}{-}                                                                  & \myx                                                                   & \mycheck                                                      & \multicolumn{1}{c}{\myx}                                                            & \multicolumn{1}{r}{$21,237$}                                        \\
MEE              & $13,000$                           & $17,642$                            & -                             & $13,548$                           & -                            & $190,592$                                                                              & \mycheck                                                               & \mycheck                                                      & $13,548$                                                                            & \myx                                                                \\
GENEVA           & \multicolumn{1}{r}{-}              & $7,505$                             & -                             & $12,269$                           & -                            & $36,390$                                                                               & \mycheck                                                               & \myx                                                          & $8,544$                                                                             & \multicolumn{1}{r}{$3,725$}                                         \\
\midrule
\ourdata         & $4,480$                            & $98,591$                            & \multicolumn{1}{r}{$107,507$} & $290,613$                          & \multicolumn{1}{r}{$83,645$} & $129,126$                                                                              & \mycheck                                                               & \mycheck                                                      & $116,024$                                                                           & \multicolumn{1}{r}{$174,589$} \\ \bottomrule                                     
\end{tabular}
}
\end{adjustbox}
\caption{Statistics of \ourdata compared to existing widely-used EAE datasets. ``Doc.'' is short for ``Document'' and ``Arg.'' is short for ``Argument''. ``-'' denotes not applicable due to lack of document structure or corresponding annotations. ``Fine-grained Event'' means annotating all the events rather than only one topic event for a document. ``Doc. Level'' means annotating arguments within the whole document rather than only the sentence containing the trigger. For multilingual datasets, we only compare with its English subset.}
\label{tab:stat}
\end{table*}

\subsection{Event Argument Annotation}
\label{sec:arg_annotation}
\looseness=-1 Based on the event detection annotations of \maven and event coreferences of \mavenere, we conduct event argument annotations. For multiple coreferent event mentions (triggers), only one of them is displayed during annotation to reduce annotation overhead. Once the annotator selects an event trigger, the corresponding argument roles for its event type are displayed on the annotation interface, along with definitions and examples. This ensures that annotators do not have to memorize the lengthy event schema or frequently refer to the annotation guidelines. To annotate an event argument, annotators can either choose an entity from the whole document or select a continuous textual span; once an entity mention is selected, all of its coreferent entity mentions are automatically selected. Annotators have the option to report errors in the event type annotation of a trigger, which allows for the discarding of that trigger. In the annotation process, approximately $4$\% of triggers are discarded. 

We employ $202$ annotators, including $71$ senior annotators selected during annotation training and $33$ experts with rich annotation experiences. The experts are recommended by the commercial data annotation companies we employed, and they have undergone at least ten data annotation projects and led at least one. The annotation is divided into three phases. Each document is first annotated by an ordinary annotator, and then modified by a senior annotator. Finally, an expert will check whether the annotation accuracy reaches $90$\%. If not, the document's annotation will be returned to the second phase. To measure data quality, we randomly sample $100$ documents and conduct the three-phrase annotation for them twice with different annotator groups. The Fleiss' kappa is $68.6$\%, which indicates a satisfactory level of annotation agreement. More annotation details are shown in \cref{app:annotation}.

\section{Data Analysis}
To provide intuitive descriptions for \ourdata, we conduct data analyses in this section.
\subsection{Data Statistics}

\cref{tab:stat} shows the main statistics of \ourdata compared with various existing EAE datasets. \Cref{app:split_stat} further shows the statistics of different splits. We can observe that \ourdata has two advantages: (1) \ourdata has the largest data scale, surpassing previous datasets by several times. This ensures that even for long-tail event types, \ourdata has sufficient data to fully train and stably evaluate EAE models. (2) The exhaustive annotation of \ourdata makes it the only dataset that covers all settings of EAE task. \ourdata includes complete annotations of entity and event coreference and annotates both entity and non-entity arguments for all fine-grained events at the document level. This allows \ourdata to support the evaluation of all variants of EAE methods and the development of comprehensive event understanding applications.

\subsection{Data Distribution}

\begin{figure}[t!]
    \centering
    \includegraphics[width=0.99\linewidth]{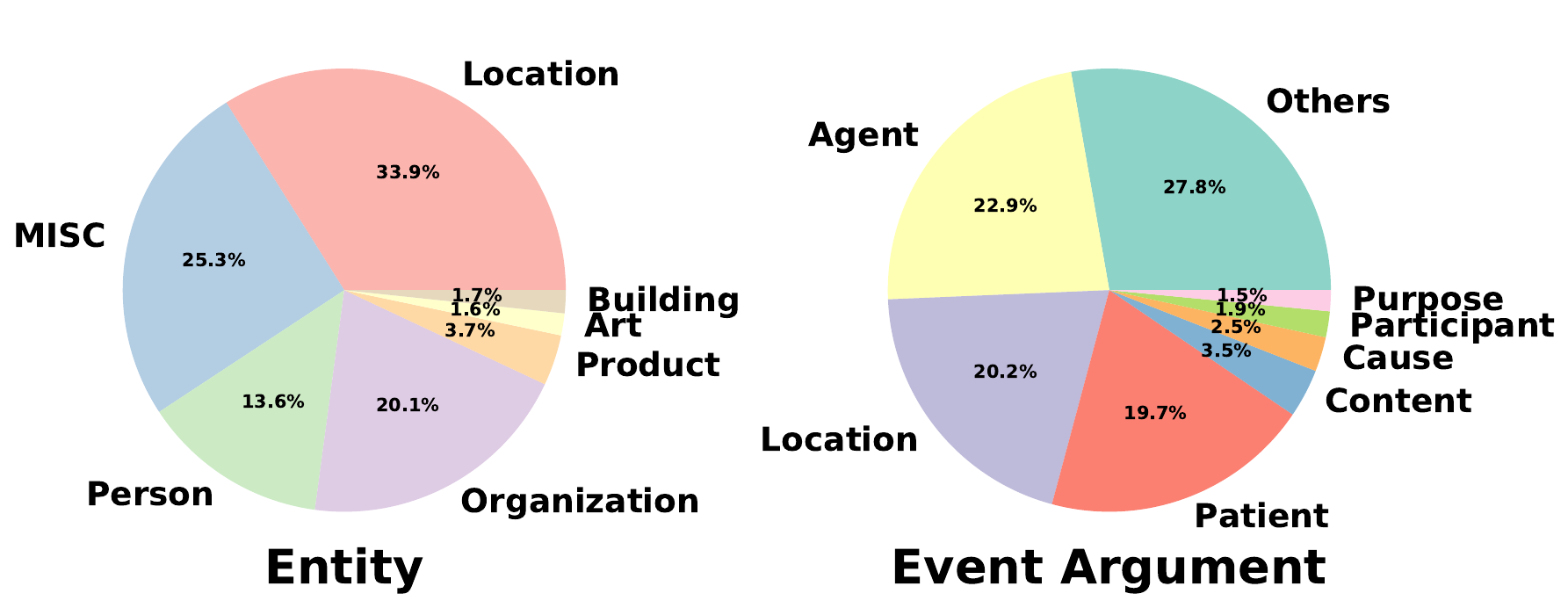}
    \caption{\ourdata entity and event argument distributions. For clarity, only the top event argument roles are shown and the others are summed up in ``Others''. }
    \label{fig:type_dist}
\end{figure}

We present the distributions of the annotated entity and event arguments of \ourdata in \cref{fig:type_dist}. Argument roles with the same name across different event types are merged for presentation clarity. We observe that: (1) The distribution of entity types is generally similar to that of Few-NERD~\citep{ding-etal-2021-nerd}, demonstrating sufficient diversity. (2) The three most frequent basic argument roles (\texttt{Agent}, \texttt{Patient}, and \texttt{Location}) account for over $60$\% of event arguments. This highlights their ubiquity and encourages knowledge transfer among different event types in EAE methods. (3) Event arguments exhibit a highly long-tailed distribution. The $136$ argument roles counted as ``Others'', each constituting less than $1.5$\%, collectively accounts for $27.8$\% of event arguments. The long-tailed distribution of \ourdata poses a significant challenge to model generalizability.

\subsection{Trigger-argument Distance}
We analyze the distribution of trigger-argument distances in \cref{fig:distance}. For events with multiple coreferent triggers and entity arguments with multiple entity mentions, the distance is calculated between the nearest trigger-argument pairs. The overall average trigger-argument distance is $37.8$. From \cref{fig:distance}, we observe that while the majority of event arguments are located near their triggers, which is natural for human writing, a substantial number of arguments are situated far from their triggers, with the furthest exceeding $800$ words. This data characteristic challenges the ability of EAE methods to capture long-distance dependencies.

\section{Experiment}

\subsection{Experimental Setup}
\label{sec:exp_setup}

\paragraph{Models} 
To assess the challenge of \ourdata, we evaluate multiple advanced methods. For fine-tuned EAE models, we implement several state-of-the-art models, including \textbf{DMBERT}~\citep{wang-etal-2019-hmeae}, \textbf{CLEVE}~\citep{wang-etal-2021-cleve}, \textbf{BERT+CRF}~\citep{wang-etal-2020-maven}, \textbf{EEQA}~\citep{li-etal-2020-event}, \textbf{Text2Event}~\citep{lu-etal-2021-text2event}, and \textbf{PAIE}~\citep{paie}. These methods cover all the mainstream EAE modeling paradigms~\citep{peng-etal-2023-devil}. Their detailed descriptions and implementations are introduced in \cref{app:fine-tune-details}. Recent works have achieved improvements in low-resource event extraction settings~\citep{liu-etal-2023-document,ma-etal-2023-large}. Considering our experiments prioritize understanding the challenge of large-scale \ourdata over conducting comprehensive evaluations for existing models, we choose not to include those low-resource methods.

We also evaluate large language models (LLMs) with in-context learning on \ourdata. Specifically, we select two advanced LLMs, \textbf{GPT-3.5}~\citep{chatgpt} and \textbf{GPT-4}~\citep{openai2023gpt}, and evaluate them with $2$-shot in-context learning. Here $2$-shot means using full annotations of two documents as demonstrations. Considering time and cost constraints, we sample $50$ documents from the test set for experimentation. We employ the gold trigger evaluation approach~\citep{peng-etal-2023-devil} to directly assess their EAE performance.

\begin{figure}[t!]
    \centering
    \includegraphics[width=0.88\linewidth]{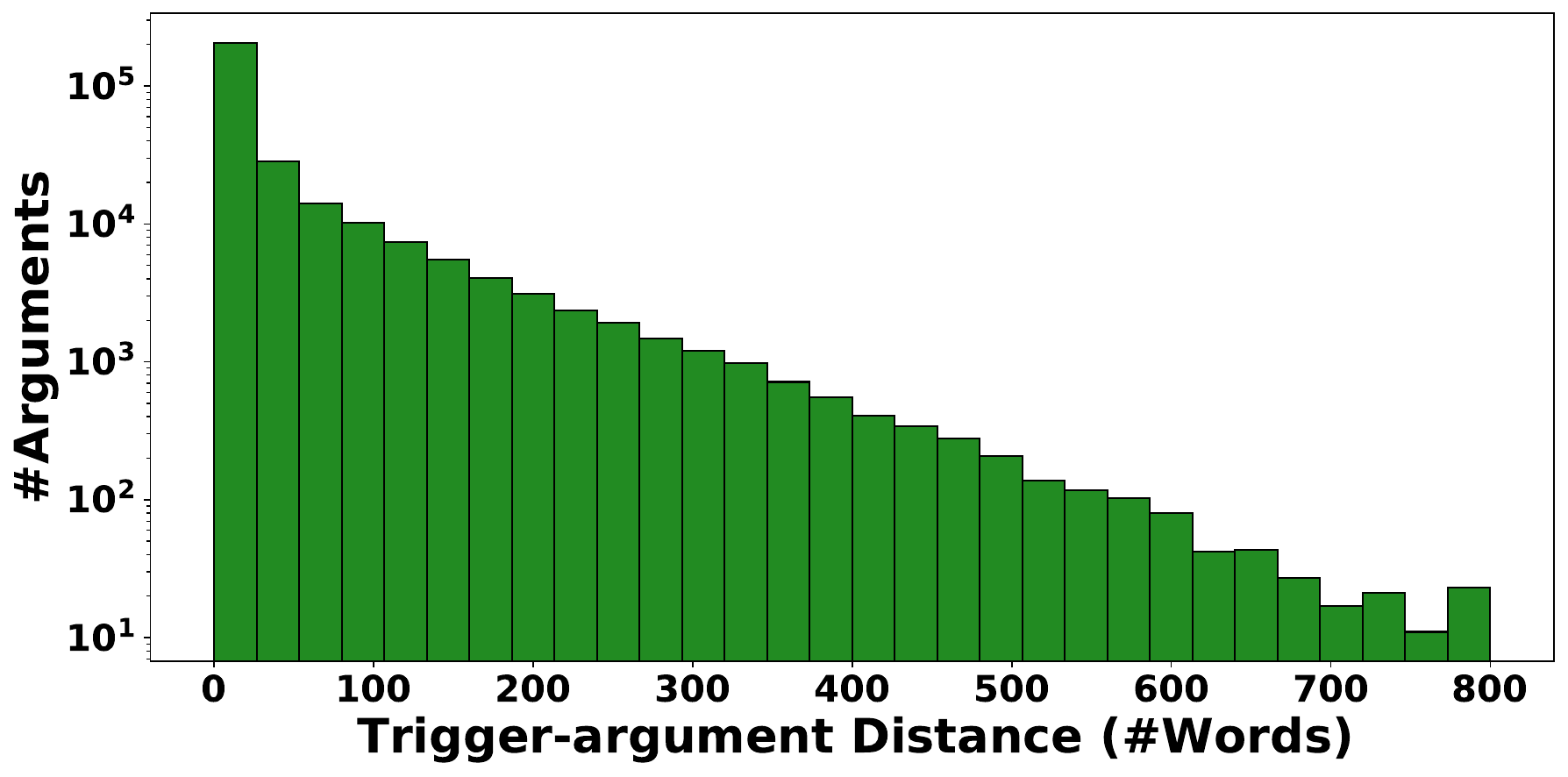}
    \caption{Distribution of distances between triggers and arguments in \ourdata.}
    \label{fig:distance}
\end{figure}

\begin{table*}[t!]
    \centering
    \begin{adjustbox}{max width=1\linewidth}
{
    \begin{tabular}{l|c|rrrr|rrrr|rrrr}
    \toprule
    \multirow{2}{*}{\textbf{Model}} & \multirow{2}{*}{\textbf{\#Params}}& \multicolumn{4}{c|}{\textbf{Mention Level}} & \multicolumn{4}{c|}{\textbf{Entity Coref Level}} & \multicolumn{4}{c}{\textbf{Event Coref Level}} \\
    & & \multicolumn{1}{c}{\textbf{P}}& \multicolumn{1}{c}{\textbf{R}} & \multicolumn{1}{c}{\textbf{F1}}& \multicolumn{1}{c|}{\textbf{EM}}& \multicolumn{1}{c}{\textbf{P}}& \multicolumn{1}{c}{\textbf{R}} & \multicolumn{1}{c}{\textbf{F1}}& \multicolumn{1}{c|}{\textbf{EM}}& \multicolumn{1}{c}{\textbf{P}}& \multicolumn{1}{c}{\textbf{R}} & \multicolumn{1}{c}{\textbf{F1}}& \multicolumn{1}{c}{\textbf{EM}}\\
    \midrule
    DMBERT & $110$M &$19.7$&$19.7$&$19.7$&$19.5$&$12.5$&$12.4$&$12.4$&$12.3$&$11.8$&$11.8$&$11.8$&$11.6$ \\
    CLEVE & $355$M & $22.1$&$22.1$&$22.1$&$22.0$&$13.2$&$13.2$&$13.2$&$13.0$&$12.3$&$12.2$&$12.2$&$12.1$ \\
    BERT+CRF & $110$M & $31.7$&$31.4$&$30.9$&$27.0$&$33.5$&$32.8$&$32.2$&$27.1$&$32.3$&$31.8$&$31.2$&$26.3$ \\
    EEQA & $110$M & $21.4$&$19.5$&$19.6$&$15.8$&$24.5$&$22.9$&$22.8$&$18.8$&$23.7$&$22.2$&$22.1$&$18.1$ \\
    Text2Event & $770$M & $12.9$&$12.9$&$12.7$&$11.3$&$12.5$&$12.4$&$12.1$&$10.4$&$10.8$&$10.7$&$10.5$&$9.0$ \\
    PAIE & $406$M & $\mathbf{37.2}$&$\mathbf{36.2}$&$\mathbf{35.6}$&$\mathbf{30.3}$&$\mathbf{42.3}$&$\mathbf{41.1}$&$\mathbf{40.5}$&$\mathbf{34.4}$&$\mathbf{42.1}$&$\mathbf{41.0}$&$\mathbf{40.3}$&$\mathbf{34.3}$ \\
    \bottomrule
    \end{tabular}
}
    \end{adjustbox}
    \caption{Experimental results (\%) of existing state-of-the-art fine-tuned EAE models on \ourdata.}
    \label{tab:finetuning_result}
\end{table*}

\paragraph{Evaluation Metric}

Considering that \ourdata covers non-entity argument annotations, traditional evaluation metrics~\citep{peng-etal-2023-devil} designed only for entity arguments are no longer applicable. By taking each argument role as a question to the document, we propose to view EAE as a \textbf{multi-answer question answering} task\footnote{A single role may correspond to multiple argument spans (answers).} and adopt its evaluation metrics~\citep{rajpurkar-etal-2016-squad,amouyal-qampari-2022, yao-etal-2023-korc}, including \textbf{bag-of-words F1} and \textbf{exact match (EM)}. 

Conventional evaluation calculates the micro average over all the entity and event mentions, which we dub it as \textbf{mention-level} evaluation. Considering that real-world applications only require the accurate prediction for one of all the coreferent mentions, we propose to consider entity~\citep{wikievents} and event coreference in evaluation. Specifically, for \textbf{entity coreference level} evaluation, an entity argument is considered as predicted correctly if one of its mentions is predicted correctly. For \textbf{event coreference level} evaluation, an argument is considered as predicted correctly if it is predicted correctly for one of the coreferent triggers.

\subsection{Experiment Results of Fine-tuned Models}

The results of fine-tuned EAE models are shown in \cref{tab:finetuning_result}, and we have the following observations: 

(1) Existing state-of-the-art EAE models exhibit moderate performance on \ourdata, which is significantly worse than their results on existing datasets~\citep{peng-etal-2023-devil}. This indicates that \ourdata is challenging and there is a need for increased efforts in developing practical event understanding models.
(2) The BERT+CRF and PAIE models exhibit the best performance, potentially attributable to their ability to model rich interactions between different event arguments. 
(3) The previous top-performing classification-based models (DMBERT and CLEVE)~\citep{peng-etal-2023-devil} perform poorly on \ourdata, which is due to their inability to handle non-entity arguments. Therefore, future research necessitates more flexible approaches to tackle the complex and real-world scenario in \ourdata.
(4) Text2Event notably underperforms. This is potentially due to the intensive annotations of \ourdata, i.e., a high volume of events and argument annotations within a single document, making generating all events and arguments at once difficult. It indicates that generating complex structured outputs remains a major challenge for generation models~\citep{peng2023specification}, requiring further exploration.
\subsection{Experiment Results of LLMs}

\begin{table*}[t!]
    \centering
    \small
    \begin{tabular}{l|rrrr|rrrr|rrrr}
    \toprule
    \multirow{2}{*}{\textbf{Model}} & \multicolumn{4}{c|}{\textbf{Mention Level}} & \multicolumn{4}{c|}{\textbf{Entity Coref Level}} & \multicolumn{4}{c}{\textbf{Event Coref Level}} \\
    & \multicolumn{1}{c}{\textbf{P}}& \multicolumn{1}{c}{\textbf{R}} & \multicolumn{1}{c}{\textbf{F1}}& \multicolumn{1}{c|}{\textbf{EM}}& \multicolumn{1}{c}{\textbf{P}}& \multicolumn{1}{c}{\textbf{R}} & \multicolumn{1}{c}{\textbf{F1}}& \multicolumn{1}{c|}{\textbf{EM}}& \multicolumn{1}{c}{\textbf{P}}& \multicolumn{1}{c}{\textbf{R}} & \multicolumn{1}{c}{\textbf{F1}}& \multicolumn{1}{c}{\textbf{EM}}\\
    \midrule
    GPT-3.5 & $21.3$&$20.9$&$19.9$&$14.3$&$24.5$&$25.1$&$23.4$&$16.8$&$24.4$&$24.8$&$23.2$&$16.9$\\
    \quad \texttt{w/ definition} & $21.8$&$21.7$&$20.6$&$15.2$&$25.0$&$25.8$&$24.1$&$17.8$&$24.9$&$25.4$&$23.9$&$17.9$ \\
    \midrule
    GPT-4 & $25.6$&$27.2$&$25.1$&$17.9$&$28.9$&$31.7$&$28.7$&$20.2$&$27.9$&$30.5$&$27.6$&$19.5$ \\
    \quad \texttt{w/ definition} & $\mathbf{27.2}$&$\mathbf{28.7}$&$\mathbf{26.6}$&$\mathbf{19.1}$&$\mathbf{30.5}$&$\mathbf{33.3}$&$\mathbf{30.3}$&$\mathbf{21.3}$&$\mathbf{29.8}$&$\mathbf{32.3}$&$\mathbf{29.5}$&$\mathbf{21.1}$\\
    \bottomrule
    \end{tabular}
    \caption{Experimental results (\%) of LLMs with 2-shot in-context learning on \ourdata.}
    \label{tab:icl_result}
\end{table*}

The results of LLMs with in-context learning are presented in \cref{tab:icl_result}, revealing that while LLMs with in-context learning are competitive compared to some fine-tuned EAE models, they still fall significantly short of the state-of-the-art. This is consistent with previous findings, suggesting that existing LLMs with in-context learning perform notably worse on specification-heavy information extraction tasks~\citep{peng2023specification, li2023evaluating,han2023information}. The LLMs' bag-of-words F1 scores are notably higher than their exact match scores, suggesting that the LLMs' predictions tend to be free-format and do not strictly match human annotations~\citep{han2023information}.

One possible reason for the suboptimal performance is that LLMs cannot easily understand the schema from their names. Therefore, we conduct experiments with more informative prompts by incorporating definitions for each used argument role into the prompt, which are high-quality instructions used for guiding human annotators during data annotation. The results of these enhanced prompts are also shown in \cref{tab:icl_result} (\texttt{w/ definition}). There is an obvious but marginal improvement after adding definitions, possibly due to the LLMs' limitations in understanding long contexts~\citep{shaham-etal-2022-scrolls, peng2023specification, liu2023lost}.

\begin{figure}
    \centering
    \includegraphics[width=0.95\linewidth]{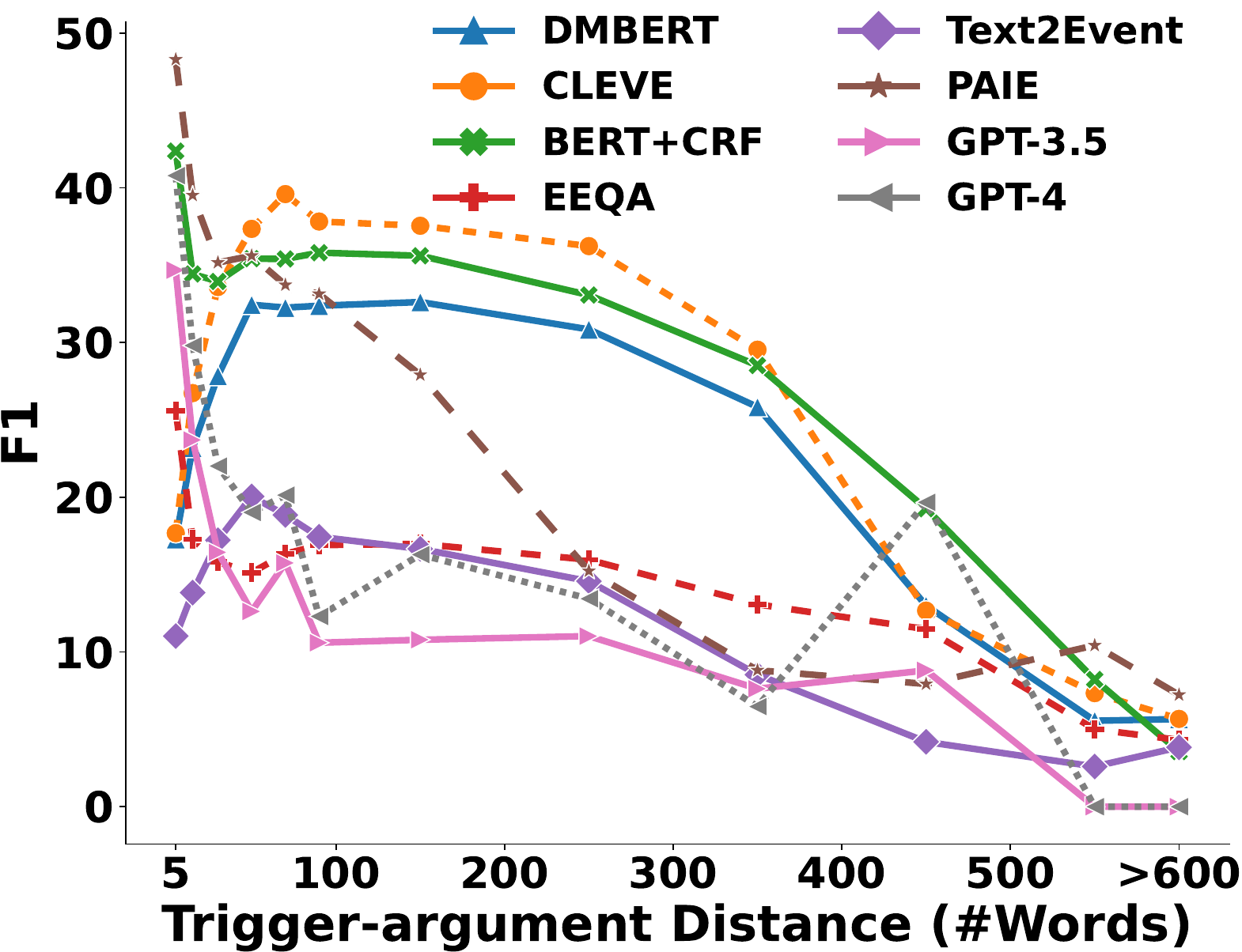}
    \caption{Mention-level F1 (\%) of models on data with varying trigger-argument distances, i.e., the number of words between an event argument and its trigger.
}
    \label{fig:f1_by_distance}
\end{figure}

\subsection{Analysis on Trigger-Argument Distance}
As shown in \cref{fig:distance}, \ourdata provides document-level annotations, covering data with varying trigger-argument distances. We conduct an analytical experiment on the impact of trigger-argument distance to model performance. Specifically, we break down the predictions and annotations in the test set by their trigger-argument distances and evaluate how the performance changes along with different distances. The experimental results are shown in \cref{fig:f1_by_distance}, which demonstrate that models perform significantly worse on samples with longer trigger-argument distances. This aligns with previous findings in document-level relation extraction regarding the distance between entity pairs~\citep{ru-learning-2021}. It suggests that modeling long-distance dependencies between triggers and arguments remains a challenge for existing EAE models. Future research can leverage \ourdata to explore advanced methods for handling long-distance trigger-argument instances.

\subsection{Analysis on Entity and Non-Entity Arguments}
\begin{table}[t]
    \centering
    \begin{tabular}{l|rr|rr}
    \toprule
    \multirow{2}{*}{\textbf{Model}} & \multicolumn{2}{c|}{\textbf{Entity}} & \multicolumn{2}{c}{\textbf{Non-Entity}} \\
     & \textbf{F1} & \textbf{EM} & \textbf{F1} & \textbf{EM} \\
    \midrule
    DMBERT & $19.7$ & $19.5$ & $-$ & $-$ \\
    CLEVE & $22.1$ & $22.0$ & $-$ & $-$ \\
    BERT+CRF & $17.8$ & $18.5$ & $19.4$ & $24.0$ \\
    EEQA & $6.2$ & $5.6$ & $17.5$ & $13.9$ \\
    Text2Event & $5.5$ & $5.2$ & $1.6$ & $1.1$ \\
    PAIE & $20.3$ & $19.2$ & $37.6$ & $30.4$  \\
    \bottomrule
    \end{tabular}
    \caption{Mention-level results (\%) of EAE models on entity and non-entity arguments. Classification-based models, e.g., DMBERT and CLEVE, are not applicable to non-entity arguments.}
    \label{tab:non_entity}
\end{table}

\ourdata provides comprehensive annotations, including both entity and non-entity arguments. We analyze the performance breakdown of investigated EAE models on these two types of arguments. The results are presented in \cref{tab:non_entity}, which reveals that EAE models generally perform better on non-entity arguments. The possible reason may be that there are more non-entity arguments in \ourdata and non-entity arguments are often presented in a looser form, making it easier for the models to learn the patterns and extract them. An exception is observed for the generation-based model Text2Event, which exhibits poorer performance on non-entity arguments. This may be because non-entity arguments are typically longer, which are harder to generate at once. It suggests that further exploration is needed to investigate how to effectively handle EAE with generation methods.

\subsection{Error Analysis}

\cref{tab:finetuning_result} shows that all the baselines can only achieve moderate performance on \ourdata, which demonstrates the challenge of our new dataset. To better understand the challenge of \ourdata and provide insights for future improvements, we conduct an error analysis for the top-performing PAIE~\citep{paie} model. One of the authors manually checked the erroneous predictions on the arguments of $50$ randomly sampled event triggers and categorized the errors into four categories: (1) \textbf{False Negative} ($52.2$\%), which means PAIE falsely ignores some event arguments; (2) \textbf{False Positive} ($12.5$\%), which means the model falsely identify non-argument entities/spans as event arguments; (3) \textbf{Span Error} ($35.3$\%), which means the model partially corrects predicts an event argument but fails to precisely predict its exact textual span; (4) \textbf{Classification Error} ($0.02$\%), which means the model correctly predicts the span of an event argument but misclassifies its argument role. We can see that similar to MAVEN~\citep{wang-etal-2020-maven}, the majority of errors are still identification errors, which means that identifying event arguments from numerous entities/spans is much more challenging than determining specific argument roles.

\section{Future Event Prediction Demonstration}
\label{sec:future_pred}
\ourdata, in conjunction with \maven and \mavenere, creates the first all-in-one event understanding benchmark, which covers the full process of ED, EAE, and ERE. Beyond serving as an evaluation benchmark for these tasks, an all-in-one event dataset naturally enables a variety of event-based applications, especially considering the recent advances brought by LLMs. Here we preliminarily explore an application case, future event prediction, as a demonstration.
 
Predicting future events based on causality can help decision-making, which is of self-evident importance. Therefore, since the early script learning~\citep{schank1975scripts,mooney1985learning}, future event prediction has continually attracted research interest~\citep{chambers-jurafsky-2008-unsupervised,jans2012skip,granroth2016happens,hu2017happens,chaturvedi-etal-2017-story,li2018constructing,lee-goldwasser-2019-multi,10.1145/3450287}. However, due to the lack of high-quality event resources, the evaluation of future event prediction often compromises by merely predicting verbs and subjects~\citep{chambers-etal-2014-dense}, predicting according to textual order~\citep{jans2012skip}, or selecting story endings~\citep{mostafazadeh-etal-2016-corpus,chaturvedi-etal-2017-story}. The MAVEN series of datasets, with annotations of complete event structures and rich causal relations, may aid in predicting future events in real-world scenarios.
\begin{table}[t!]
\centering
\begin{tabular}{lrr}
\toprule
\textbf{Model} & \multicolumn{1}{c}{\textbf{Reasonable (\%)}} & \multicolumn{1}{c}{\textbf{Matched (\%)}} \\
\midrule
GPT-3.5        & $92.7$                                       & $7.8$                                     \\
GPT-4          & $95.2$                                       & $12.2$   \\
\bottomrule                                
\end{tabular}
\caption{Future event prediction results (\%), averaged over $2$ evaluators and $3$ prompts. \textbf{Reasonable} denotes the rate of predictions judged as reasonable to happen next. \textbf{Matched} denotes the rate of predictions matched with the actual future events.}
\label{tab:pred}
\end{table}

\paragraph{Experiment Setup}
We sample $100$ event chains, each consisting of $3$-$5$ events, from the training and validation sets. In each chain, preceding events cause the subsequent ones. Events are described in a structured JSON format, containing event type, event trigger, and event arguments. For each event chain, we hold out the last event and input the remaining incomplete chain into two proprietary LLMs, GPT-3.5 and GPT-4~\citep{openai2023gpt}, requiring them to predict the next occurring event. These LLMs are prompted with detailed task instructions and $5$ demonstration event chains. To minimize the influence of the demonstrations, predictions are made independently three times under different demonstrations. More experimental details are shown in \cref{app:pred}. We employ manual evaluation, with two experts engaged to judge (1) whether the prediction is reasonable, and (2) whether the prediction matches the actual future event.

\paragraph{Experimental Results}

\looseness=-1 Experimental results are shown in \cref{tab:pred}. From these, we can see that the powerful LLMs can produce highly reasonable event predictions. However, their predictions seldom align with the actual future, making them not directly helpful. These observations suggest that using LLMs for future event prediction is promising, but there remain topics to explore on how to build practical future event prediction systems with LLMs. For instance, using retrieval-augmented methods may help LLMs access more timely evidence when making future predictions. As a preliminary attempt, the experiments demonstrate how our all-in-one event understanding dataset can assist in conveniently building and evaluating event-based applications. We hope that future works can explore using the MAVEN series datasets to build diverse applications.
\section{Related Work}
\paragraph{Event Argument Extraction Datasets}
Since the early MUC datasets~\citep{grishman-sundheim-1996-message}, event argument extraction (EAE) as a part of event extraction has received widespread attention. To reduce task complexity and provide standardized extraction results, the ACE datasets~\citep{doddington2004automatic} are designed with a schema covering $33$ event types, limiting event argument annotation to entities within the same sentence as the trigger. ACE 2005~\citep{walker2006ace} has been the most widely used dataset for a long time, and the practice of ACE has been broadly adopted. Rich ERE~\citep{song-etal-2015-light} expands ACE schema to $38$ event types and constructs the TAC KBP datasets~\citep{ellis2014overview,ellis2015overview,ellis2016overview,getman2017overview}. MEE~\citep{pouran-ben-veyseh-etal-2022-mee} follows the ACE schema to build a multilingual dataset. With the advancement of NLP methods, some works break some of the constraints of ACE task definition to construct more practical datasets. RAMS~\citep{rams}, WikiEvents~\citep{wikievents}, and DocEE~\citep{tong-etal-2022-docee} extends the annotation scope to the whole documents. However, RAMS and DocEE only annotate one topic event per document, ignoring fine-grained events within documents. MAVEN~\citep{wang-etal-2020-maven} and GENEVA~\citep{parekh-etal-2023-geneva} both construct high-coverage general event schemata with over $100$ event types. MAVEN supports only event detection. GENEVA extends event arguments to cover non-entity spans but focuses on testing the generalizability rather than developing practical EAE methods. Its data are repurposed from FrameNet~\citep{baker1998berkeley} examples, which are individual sentences without document structure. \ourdata meticulously designs $612$ unique argument roles for MAVEN schema and conducts large-scale exhaustive annotation, which annotates both entity and non-entity arguments for fine-grained events at the document level.

\paragraph{Event Argument Extraction Methods}
\looseness=-1 Traditional EAE methods primarily involve (1) Classification-based methods~\citep{chen2015event, chen-etal-2017-automatically, sha-jointly-2018, wadden-dygie-2019, wang-etal-2019-hmeae, lin-etal-2020-joint, wang-etal-2021-cleve, zhou-mao-2022-document}: employing text encoders like CNN~\citep{cnn} and BERT~\citep{bert}, followed by an information aggregator, such as dynamic multi-pooling mechanism~\citep{chen2015event}, to obtain role-specific representations for classification. (2) Sequence labeling methods~\citep{nguyen-etal-2016-joint-event, yang-leveraging-2017, nguyen-etal-2021-crosslingual, peng-etal-2023-devil}: mainly adopting the conditional random field (CRF)~\citep{crf} as the output layer to model structured dependencies between different arguments. Recently, increasing attention has been paid to transforming EAE into a question-answering task, transferring question-answering capabilities to boost EAE~\citep{liu2020event, du2020event, li-etal-2020-event, paie, lu-etal-2023-event, nguyen-etal-2023-contextualized}. Additionally, some research focuses on using generation models to directly generate structured outputs containing events and their arguments~\citep{lu-etal-2021-text2event, wikievents,  lu-etal-2022-unified, ren-etal-2023-retrieve, you-etal-2022-eventgraph, you-etal-2023-jseegraph,hsu2022degree, hsu-etal-2023-ampere, zhang-etal-2023-overlap,ren-etal-2023-retrieve,liu-etal-2023-document,ma-etal-2023-large}, which has been becoming increasingly important with the advance of large language models. 

\section{Conclusion and Future Work}
We introduce \ourdata, an event argument extraction dataset with comprehensive schema, large data scale, and exhaustive annotation. Experiments indicate that \ourdata is quite challenging for both fine-tuned EAE models and proprietary large language models. Together with \maven and \mavenere, \ourdata completes an all-in-one dataset covering the entire process of event understanding. An application case of future event prediction demonstrates how an all-in-one dataset can enable broad event-based applications. In the future, we will explore constructing multilingual resources under this framework and developing practical EAE methods with \ourdata.

\section*{Acknowledgements}
This work is supported by a grant from the Institute for Guo Qiang, Tsinghua University (2019GQB0003). We thank all the annotators for their efforts and the anonymous reviewers for their valuable comments.

\section*{Limitations}

\looseness=-1 (1) \ourdata currently includes only English corpus, which limits its potential applications and coverage for diverse linguistic phenomena. In future work, we will try to support more languages under our framework and we also encourage community efforts in developing multilingual event understanding benchmarks. (2) \ourdata, along with \maven~\citep{wang-etal-2020-maven} and \mavenere~\citep{wang2022maven}, exclusively supports mainstream event understanding tasks. However, these datasets do not cover more broad event-related tasks such as event factuality identification~\citep{qian-etal-2019-document, qian2022document} and event salience identification~\citep{liu-etal-2018-automatic}. We encourage future explorations in building more challenging and diverse tasks and applications on top of \maven data. (3) While previous research has found that LLMs perform poorly on specification-heavy tasks~\citep{peng-etal-2023-devil, han2023information, li2023evaluating} including the EAE task, there is no in-depth exploration of effective LLM-based approaches addressing the EAE task in this paper. We leave the exploration of how to better leverage LLMs for EAE tasks in future work.

\section*{Ethical Considerations}
In this section, we discuss the ethical considerations of this work: (1) \textbf{Intellectual property.} The \maven dataset
is released under the CC BY-SA 4.0 license\footnote{\url{https://creativecommons.org/licenses/by-sa/4.0/}}. The \mavenere is shared under GPLv3\footnote{\url{https://www.gnu.org/licenses/gpl-3.0.html}} license and the original Wikipedia corpus is shared under the CC BY-SA 3.0 license\footnote{\url{https://creativecommons.org/licenses/by-sa/3.0/}}. The usage of these data in this work strictly adheres to the corresponding licenses and intended use. (2) \textbf{Intended use.} \ourdata is an event argument extraction dataset. Researchers and practitioners can utilize \ourdata to train and evaluate models for event argument extraction, thereby advancing the field of event understanding. (3) \textbf{Potential risk control.} \ourdata is constructed based on publicly available data. We believe that the underlying public data has been adequately desensitized and anonymized. The event argument annotation does not involve judgments about social issues and thus we believe \ourdata will not involve additional risks. To avoid unfair comparisons caused by mismatched evaluation implementations~\citep{peng-etal-2023-devil} and potential cheating behaviors, the event argument annotations of \ourdata test set will not be publicly released. Instead, following previous works~\citep{rajpurkar-etal-2016-squad,wang-etal-2020-maven,wang2022maven}, we will maintain an online judgment system with a leaderboard, allowing users to submit predictions and obtain evaluation results. (4) \textbf{Worker Treatments} are discussed in \cref{app:annotation_cood}.

\bibliography{custom} 

\clearpage
\appendix

\section*{Appendices}
\section{Data Collection Details}
\label{app:annotation}
\subsection{Annotation Instruction}
\label{app:annotation_struction}

\begin{figure*}[t!]
    \centering
    \includegraphics[width=0.99\linewidth]{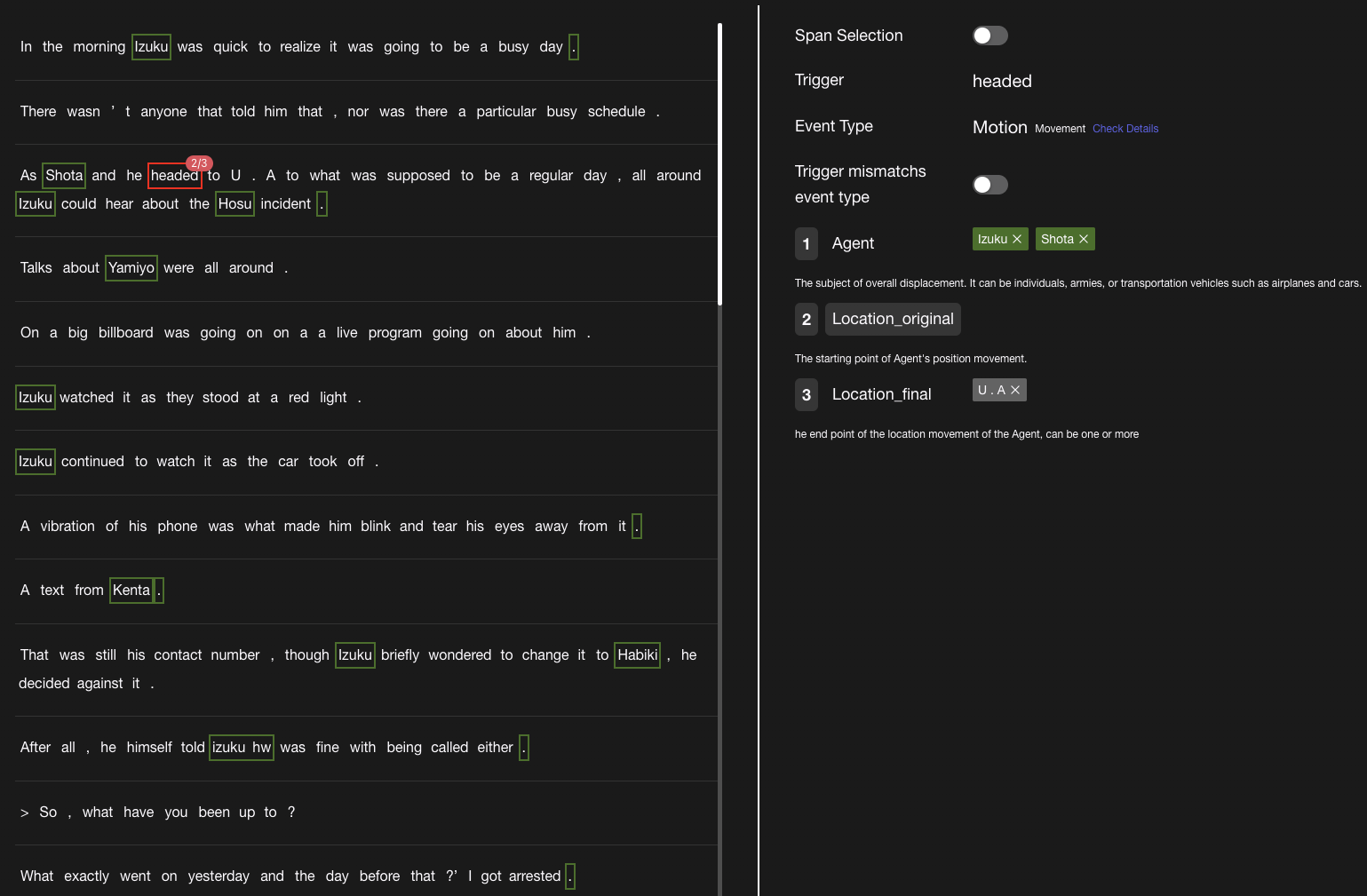}
    \caption{Screenshot for the annotation platform. The trigger ``headed'' is selected for annotation (in the right panel) and entities are highlighted in green as the options for annotating event arguments.}
    \label{fig:platform}
\end{figure*}

As introduced in \cref{sec:schema}, we create a detailed event schema for both defining the task and instructing the annotators. We present the annotation instructions for the event type \texttt{Incident} in \cref{tab:annotation_instruction}, including its argument schema and annotation examples. The overall event schema is released along with the dataset. To support the highly customized annotation process designed for us, we developed a new online annotation platform. A screenshot for the annotation platform is shown in \cref{fig:platform} to help understand the annotation operations.

\begin{table*}[ht]

    \centering
    \small
    \begin{tabular}{p{\linewidth}}
        \toprule

            \vspace{-2mm}
		\textbf{[Incident] Accident, unfortunate event}\\
        \vspace{-1mm}
        \textbf{Event Arguments:}\\
        \vspace{-1mm}
 		1. \colorbox{yellow}{Participant}: Entities involved in the accident (individuals, institutions, organizations, and even trains, ships, etc.). They can be the ones causing the accident or the ones affected by it. Similar to the combination of Agent and Patient in previous events, but due to the difficulty in distinguishing between Agent and Patient in accidents, they are uniformly labeled as Participants. \\
 		\vspace{-1mm}
 		2. \colorbox{cyan}{Location}: The location or position where the incident occurred. If the incident involves multiple locations during the process, they should be marked separately. \\
 		\vspace{-1mm}
 		3. \colorbox{red}{Content}: In general, only one annotation is needed, which accurately indicates the content and type of the accident.\\
 		\vspace{-1mm}
 		4. \colorbox{brown}{Loss}: The losses caused by accidents can include the number of deaths and injuries, property damage, and so on.\\
 		\vspace{-1mm}
        \textbf{Annotation Examples:}\\
         		\vspace{-1mm}
 		1. British losses were confined to \colorbox{brown}{a single man wounded} by an \textcolor{red}{\textbf{accident}} \colorbox{cyan}{aboard ``Crescent''}.\\
 		 \vspace{-1mm}
 		2. On 6 June 1982, during the Falkland’s war, \colorbox{yellow}{the British Royal Navy type 42 destroyer} engaged and destroyed a [British army gazelle helicopter, serial number ``XX377'']\colorbox{yellow}{Participant}+\colorbox{brown}{Loss}, in a \colorbox{red}{friendly fire} \textcolor{red}{\textbf{incident}}, \colorbox{brown}{killing all four occupants}.\\ \bottomrule
    \end{tabular}
        \caption{Example annotation instructions for event type \texttt{Incident}. Different argument roles are denoted by different background colors. \textcolor{red}{\textbf{Triggers}} are bolded in red.}
    \label{tab:annotation_instruction}
\end{table*}

\subsection{Annotation Coordination}
\label{app:annotation_cood}
We employ annotators (including senior annotators and expert annotators) from multiple commercial data annotation companies. $61$\% of them are female and $39$\% of them are male. All annotators hold undergraduate degrees. Annotators for entity and event argument annotation have no overlap since we cooperated with different companies for the two annotation tasks. The experts involved in schema creation are invited by the authors through personal connections. All the workers are fairly paid with agreed salaries and workloads. All employment is under contract and in compliance with local regulations. The overall annotation cost, including annotating entities and event arguments as well as developing and maintaining annotation platforms, is about $85,000$ USD.

\begin{table*}[t!]
\centering
\small
    \begin{adjustbox}{max width=1\linewidth}
{
\begin{tabular}{lrrrrrrrrrr}	
\toprule
\multicolumn{1}{c}{\textbf{Dataset}} & \multicolumn{1}{c}{\textbf{\#Doc.}} & \multicolumn{1}{c}{\textbf{\#Event}} & \multicolumn{1}{c}{\textbf{\#Trigger}} & \multicolumn{1}{c}{\textbf{\#Arg.}} & \multicolumn{1}{c}{\textbf{\#Entity}} & \multicolumn{1}{c}{\textbf{\begin{tabular}[c]{@{}c@{}}\#Entity\\ Mention\end{tabular}}} & \textbf{\begin{tabular}[c]{@{}c@{}}Entity \\ Arg.\end{tabular}} & \textbf{\begin{tabular}[c]{@{}c@{}}Non-Entity \\ Arg.\end{tabular}} \\
\midrule 
Train         & $2,913$                              & $64,923$                             & \multicolumn{1}{r}{$70,775$}   & $190,479$                            & \multicolumn{1}{r}{$55,421$} & $86,969$                                                                                                                                      & $76,882$                                                                             & $113,597$                                                                \\
Dev            & $710$                           & $15,556$                            & $16,996$                             & $46,458$                          & $12,927$                            & $18,806$                                                                 & $18,040$               & $28,418$                                 \\
Test       & $857$                              & $18,112$                             & $19,736$                             & $53,676$                            & \multicolumn{1}{r}{$15,297$} & $23,351$                                                                                                                                    & $21,102$                                                                             & $32,574$                                                                \\ \bottomrule                                     
\end{tabular}
}
\end{adjustbox}
\caption{Statistics of the data splits of \ourdata. ``Doc.'' is short for ``Document'' and ``Arg.'' is short for ``Argument''.}
\label{tab:split_stat}
\end{table*}

\section{Additional Data Statistics}
\subsection{Data Split Statistics}
\label{app:split_stat}
The detailed statistics of different data splits of \ourdata are shown in \cref{tab:split_stat}.

\subsection{Differences with Predecessors}
\ourdata inherits the efforts of previous works \maven~\citep{wang-etal-2020-maven} and \mavenere~\citep{wang2022maven}. \maven supports the event detection task by annotating event triggers and event types, along with a preliminary version of event coreferences. \mavenere supports the event relation extraction task by annotating event coreference, temporal, causal, and hierarchical relations. \ourdata completes the all-in-one event understanding dataset by adding the annotations of event arguments, which supports the event argument extraction task. The construction of \mavenere and \ourdata involves fixing or ignoring the erroneous and ambiguous annotations of event triggers and coreference clusters in \maven, which results in minor statistical differences shown in \cref{tab:difference}.

\begin{table}[t]
    \centering
    \small
\begin{tabular}{lrr}
\toprule
\textbf{} & \multicolumn{1}{c}{\textbf{\#Trigger}} & \multicolumn{1}{c}{\textbf{\#Coreference Cluster}} \\
\midrule 
\maven    & $118,732$                             & $111,611$                                         \\
\mavenere & $112,276$                             & $103,193$                                         \\
\ourdata  & $107,507$                             & $98,591$                   \\                      
\bottomrule
\end{tabular}
    \caption{Statistical differences between \ourdata and predecessors in number of event triggers and coreference clusters.}
    \label{tab:difference}
\end{table}
\section{EAE Experimental Details}

\subsection{Fine-tuning Implementation Details}
\label{app:fine-tune-details}

Here we provide brief descriptions of the fine-tuning-based models involved in our experiments. (1) \textbf{DMBERT}~\citep{wang-etal-2019-hmeae} utilizes BERT~\cite{bert} as the text encoder and a dynamic multi-pooling mechanism~\citep{chen-etal-2015-event} on top of BERT to aggregate argument-specific features and map them onto the distribution in the label space. (2) \textbf{CLEVE}~\citep{wang-etal-2021-cleve} is an event-oriented pre-trained language model, which is pre-trained using contrastive pre-training objectives on large-scale unsupervised data and their semantic structures. (3) \textbf{BERT+CRF}~\citep{wang-etal-2020-maven} is a sequence labeling model, which leverages BERT as the backbone and the conditional random field (CRF)~\citep{crf} as the output layer to model the structural dependencies of predictions. (4) \textbf{EEQA}~\citep{li-etal-2020-event} is a span prediction model, which formulates event extraction as a question-answering task and outputs start and end positions to indicate triggers and arguments. (5) \textbf{Text2Event}~\citep{lu-etal-2021-text2event} is a conditional generation model, which proposes a sequence-to-structure paradigm and generates structured outputs containing triggers and corresponding arguments with constrained decoding. (6) \textbf{PAIE}~\citep{paie} adopts prompt tuning~\citep{lester-etal-2021-power} to train two span selectors for each argument role in the provided prompt and conduct joint optimization to find optimal role-span assignments. We adopt the same backbones with their original papers for all EAE models in our experiments. We employ pipeline evaluation as suggested by~\citet{peng-etal-2023-devil}. Specifically, for PAIE, we conduct EAE experiments based on the triggers predicted by CLEVE. For the other models, the EAE experiments are based on the triggers extracted by corresponding models.

We implement the EAE models using code from the official repositories of OmniEvent~\citep{peng2023omnievent}, PAIE~\citep{paie}, and Text2Event~\citep{lu-etal-2021-text2event}. The numbers of parameters of the EAE models are shown in Table~\ref{tab:finetuning_result}. All open-source models are downloaded from the HuggingFace Transformers community~\citep{wolf-etal-2020-transformers}. Each of our fine-tuning experiments is conducted only once, on Nvidia A100 GPUs, consuming approximately 800 GPU hours in total. The hyper-parameters of the model are set based on prior experience and references from previous papers~\citep{lu-etal-2021-text2event, paie, peng2023omnievent}. All hyper-parameters are shown in \cref{tab:hyper-param}.

\begin{table*}[t]
    \centering
    \small
    \begin{tabular}{lrrrrrr}
    \toprule
    & DMBERT & CLEVE & BERT+CRF & EEQA & PAIE & Text2Event \\
    \midrule
    Learning Rate & $5 \times 10^{-5}$ & $1 \times 10^{-5}$ & $5 \times 10^{-5}$ & $5 \times 10^{-5}$ & $2 \times 10^{-5}$ & $5 \times 10^{-5}$ \\ 
    Weight Decay & $1 \times 10^{-5}$ & $1 \times 10^{-5}$ & $1 \times 10^{-5}$ & $1 \times 10^{-5}$ & $1 \times 10^{-5}$ & $1 \times 10^{-2}$ \\
    Batch Size & $32$ & $128$ & $64$ & $32$ & $16$ & $8$ \\ 
    Epoch & $6$ & $5$ & $10$ & $10$ & $-$ & $30$ \\ 
    Warmup Rate & $0.1$ & $0.1$ & $0.1$ & $0.1$ & $0$ & $0.1$ \\
    \bottomrule
    \end{tabular}
    \caption{Hyper-parameters of fine-tuning EAE models on \ourdata. PAIE utilizes $10,000$ gradient update steps to optimize the parameters.}
    \label{tab:hyper-param}
\end{table*}

\subsection{LLM Experimental Details}
We access ChatGPT and GPT-4 through the official OpenAI interfaces, namely \texttt{gpt-3.5-turbo} and \texttt{gpt-4}, respectively. The API access period spans from October 1 to October 31, 2023. The decoding sampling temperature for both models is set to 0. An example of the prompt, input, output, and ground-truth of this experiment are presented in Table~\ref{tab:app_llm_prompt}. Model outputs are automatically extracted and evaluated using the evaluation approach mentioned in \cref{sec:exp_setup}.

\begin{table*}[ht]

    \centering
    \small
    \begin{tabular}{p{\linewidth}}
        \toprule

            \vspace{-2mm}
        \cellcolor{'shallow4'} \textbf{\textsc{Prompt:}} Please extract event argument roles and the corresponding mentions for the events marked with <event> and </event> in the text, the possible roles must be chosen from the Roleset. If there are no roles for the event, place output NA.\\
        \midrule
        \vspace{-1mm}
        \cellcolor{'shallow4'} \textbf{\textsc{Input:}} The February 2010 Australian \textbf{<event>} cyberattacks \textbf{</event>} were a series of denial-of-service attacks conducted by the Anonymous online community against the Australian government in response to proposed web censorship regulations . Operation Titstorm was the name given to the cyber attacks by the perpetrators . They resulted in lapses of access to government websites on 10 and 11 February 2010 . This was accompanied by emails , faxes , and phone calls harassing government offices . The actual size of the attack and number of perpetrators involved is unknown but it was estimated that the number of systems involved ranged from the hundreds to the thousands . The amount of traffic caused disruption on multiple government websites . Australian Telecommunications Minister Stephen Conroy proposed the regulations that would mainly filter sites with pornographic content . Various groups advocating uncensored access to the Internet , along with companies like Google and Yahoo ! , object to the proposed filter . A spokesperson for Conroy said that the actions were not a legitimate form of protest and called it irresponsible . The attacks also drew criticism from other filter protest groups . The initial stage was followed by small in-person protests on 20 February that were called `` Project Freeweb '' . Roleset: [Agent, Location, Patient, Consequence]. The event trigger is: \textbf{<event>} cyberattacks \textbf{</event>}.\\
        \midrule
        \vspace{-1mm}
        \cellcolor{'shallow4'} \textbf{\textsc{GPT-3.5 Output:}} Answer: Patient: the Australian government; Consequence: lapses of access to government websites; Agent: the Anonymous online community\\
        \midrule
        \vspace{-1mm}
        \cellcolor{'shallow4'} \textbf{\textsc{GPT-4 Output:}} Answer: Agent: the Anonymous online community; Location: Australia; Patient: Australian government; Consequence: lapses of access to government websites, disruption on multiple government websites.\\
        \midrule
        \vspace{-1mm}
        \cellcolor{'shallow4'} \textbf{\textsc{Ground-truth:}} Agent: Anonymous online community; Patient: Australian government; Consequence: lapses of access to government websites on 10 and 11 February 2010; Location: Australian
\\
        \bottomrule
    \end{tabular}
        \caption{
  Example prompt, input, output, and ground-truth for the LLMs experiments. Demonstrations are omitted.
}
    \label{tab:app_llm_prompt}
\end{table*}

\section{Event Prediction Experimental Details}
\label{app:pred}
The future event prediction experiments (\cref{sec:future_pred}) were conducted in October and November, 2023. We use OpenAI API endpoints \texttt{gpt-3.5-turbo} and \texttt{gpt-4} for GPT-3.5 and GPT-4 experiments, specifically. To ensure the consistency among different runs, we set \texttt{temperature=0.0}. Detailed instructions and example input and output are shown in \cref{tab:event_pred_instruction}.

\begin{table*}[ht]

    \centering
    \small
    \begin{tabular}{p{\linewidth}}
        \toprule

            \vspace{-2mm}
        \cellcolor{'shallow3'} \textbf{\textsc{Prompt:}} You need to complete the event chain prediction task. I will provide you with an event chain consisting of multiple events, where earlier events cause the occurrence of subsequent events. You need to predict what the next most likely event in this chain will be. Events will be provided in a structured JSON format, including event type, core event trigger words, and event elements. I will first give you 5 example chains: ... Now please predict the next event for the following event chain:\\
        \midrule
        \vspace{-1mm}
        \cellcolor{'shallow3'} \textbf{\textsc{Input:}} Event No.1: \{"event type": "Attack", "event trigger": ["attack"], "event argument": \{"Agent": ["unknown assailant"], "Patient": ["Boris Nemtsov"], "Location": ["Bolshoy Moskvoretsky Bridge"]\}\} Event No.2: \{"event type": "Use\_firearm", "event trigger": ["fired"], "event argument": \{"Agent": ["unknown assailant"], "Patient": ["Nemtsov"], "Location": ["Bolshoy Moskvoretsky Bridge"], "Instrument": ["Makarov pistol"]\}\} Event No.3: \{"event type": "Bodily\_harm", "event trigger": ["hit"], "event argument": \{"Agent": ["shots"], "Patient": ["Boris Nemtsov"], "Location": ["Bolshoy Moskvoretsky Bridge"]\}\}\\
        \midrule
        \vspace{-1mm}
        \cellcolor{'shallow3'} \textbf{\textsc{GPT-3.5 Output:}} Based on the provided event chains, the next most likely event in the given event chain is: Event No.4: \{"event type": "Criminal\_investigation", "event trigger": ["indicted"], "event argument": \{"Agent": ["unknown assailant"], "Suspect": ["Boris Nemtsov"], "Incident": ["shooting on Bolshoy Moskvoretsky Bridge"], "Conclusion": ["investigation and charges"]\}\}\\
        \midrule
        \vspace{-1mm}
        \cellcolor{'shallow3'} \textbf{\textsc{GPT-4 Output:}} Event No.4: \{"event type": "Death", "event trigger": ["died"], "event argument": \{"Agent": ["Boris Nemtsov"], "Cause": ["gunshot wounds"], "Location": ["Bolshoy Moskvoretsky Bridge"]\}\}\\
        \midrule
        \vspace{-1mm}
        \cellcolor{'shallow3'} \textbf{\textsc{Ground-truth:}} Event No.4: \{"event type": "Death", "event trigger": ["died"], "event argument": \{"Agent": ["Boris Nemtsov"], "Location": ["Bolshoy Moskvoretsky Bridge"]\}\}
\\
        \bottomrule
    \end{tabular}
        \caption{
  Example prompt, input, output, and ground-truth for the future event prediction experiments. Demonstrations are omitted and the JSON strings are unformatted to avoid taking up to much space.
}
    \label{tab:event_pred_instruction}
\end{table*}
\section{More Experimental Results}
In this section, we present more experimental results of using different proportions of training data for training (\cref{sec:app_data_size}) and results on entity and non-entity arguments (\cref{sec:app_non_entity}).
\subsection{Analysis on Data Size}
\label{sec:app_data_size}
\begin{figure}[t!]
    \centering
    \includegraphics[width=0.95\linewidth]{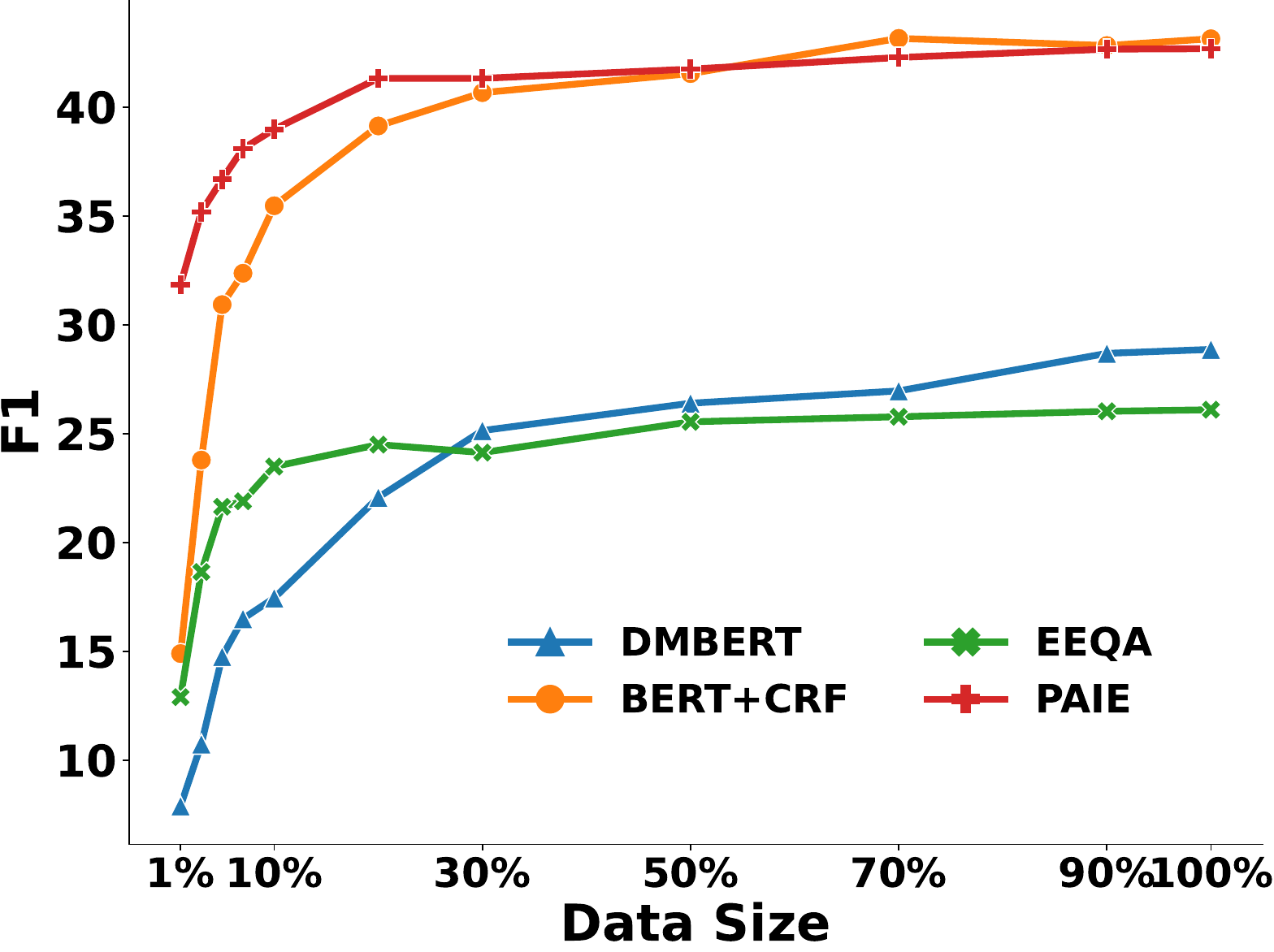}
    \caption{Mention-level F1 scores of investigated EAE models using different proportions of training data on \ourdata. This experiment adopts the gold trigger evaluation approach~\citep{peng-etal-2023-devil} and hence Text2Event is not applicable. Due to the computation limitations, CLEVE is not evaluated in this experiment.}
    \label{fig:data_size}
\end{figure}

The data volume of \ourdata significantly exceeds that of commonly used datasets. To examine the benefits of increased data scale, we train models on training data of varying sizes and observe their performance on the original test set. The experimental results are shown in \cref{fig:data_size}, which demonstrates that more training data indeed enhances model performance and allows for a comprehensive comparison of different models. The extensive data of \ourdata make it feasible to train a large language model (LLM) for general event understanding, which we leave as future work.
\cref{tab:app_data_size} shows the detailed experimental results, i.e., mention level, entity coreference level, and event coreference level.
\begin{table*}[t!]
    \centering
    \small
    \begin{tabular}{r|rrrr|rrrr|rrrr}
    \toprule
    \multirow{2}{*}{\textbf{Proportion}} & \multicolumn{4}{c|}{\textbf{Mention Level}} & \multicolumn{4}{c|}{\textbf{Entity Coref Level}} & \multicolumn{4}{c}{\textbf{Event Coref Level}} \\
    & \multicolumn{1}{c}{\textbf{P}}& \multicolumn{1}{c}{\textbf{R}} & \multicolumn{1}{c}{\textbf{F1}}& \multicolumn{1}{c|}{\textbf{EM}}& \multicolumn{1}{c}{\textbf{P}}& \multicolumn{1}{c}{\textbf{R}} & \multicolumn{1}{c}{\textbf{F1}}& \multicolumn{1}{c|}{\textbf{EM}}& \multicolumn{1}{c}{\textbf{P}}& \multicolumn{1}{c}{\textbf{R}} & \multicolumn{1}{c}{\textbf{F1}}& \multicolumn{1}{c}{\textbf{EM}}\\
    \midrule
    \multicolumn{13}{c}{\textbf{DMBERT}}\\
    \midrule
    $1\%$&$8.3$&$7.8$&$7.9$&$7.2$&$10.4$&$9.5$&$9.7$&$8.6$&$9.4$&$8.6$&$8.8$&$7.7$\\$3\%$&$11.1$&$10.7$&$10.8$&$10.1$&$11.9$&$11.2$&$11.3$&$10.3$&$10.2$&$9.6$&$9.7$&$8.8$\\$5\%$&$15.2$&$14.7$&$14.8$&$14.0$&$14.8$&$13.9$&$14.0$&$12.9$&$13.1$&$12.3$&$12.5$&$11.5$\\$7\%$&$17.0$&$16.4$&$16.5$&$15.7$&$17.9$&$16.8$&$17.1$&$15.7$&$16.4$&$15.4$&$15.6$&$14.4$\\$10\%$&$18.0$&$17.4$&$17.5$&$16.6$&$17.5$&$16.4$&$16.6$&$15.4$&$16.0$&$15.0$&$15.2$&$14.1$\\$20\%$&$22.6$&$22.0$&$22.1$&$21.2$&$21.0$&$19.8$&$20.0$&$18.6$&$19.3$&$18.2$&$18.4$&$17.1$\\$30\%$&$25.7$&$25.0$&$25.2$&$24.2$&$23.1$&$21.7$&$21.9$&$20.4$&$21.5$&$20.2$&$20.4$&$19.0$\\$50\%$&$26.9$&$26.3$&$26.4$&$25.5$&$23.9$&$22.7$&$22.9$&$21.4$&$22.1$&$21.0$&$21.2$&$19.8$\\$70\%$&$27.5$&$26.9$&$27.0$&$26.1$&$24.0$&$22.7$&$23.0$&$21.5$&$22.1$&$21.0$&$21.2$&$19.9$\\$90\%$&$29.2$&$28.6$&$28.7$&$27.8$&$24.7$&$23.5$&$23.8$&$22.2$&$22.9$&$21.8$&$22.0$&$20.6$\\
    \midrule
    \multicolumn{13}{c}{\textbf{BERT+CRF}}\\
    \midrule
    $1\%$&$16.4$&$14.8$&$14.9$&$11.6$&$21.7$&$19.6$&$19.8$&$15.4$&$21.1$&$19.1$&$19.3$&$15.1$\\$3\%$&$25.3$&$24.1$&$23.8$&$19.2$&$31.4$&$29.6$&$29.4$&$23.4$&$30.4$&$28.8$&$28.5$&$22.9$\\$5\%$&$32.5$&$31.2$&$30.9$&$25.8$&$38.8$&$36.8$&$36.5$&$29.6$&$38.0$&$36.2$&$35.9$&$29.2$\\$7\%$&$33.9$&$32.8$&$32.4$&$27.1$&$41.6$&$39.7$&$39.3$&$31.9$&$40.7$&$38.9$&$38.5$&$31.3$\\$10\%$&$36.8$&$36.0$&$35.5$&$30.0$&$43.1$&$41.8$&$41.1$&$33.9$&$42.3$&$41.1$&$40.4$&$33.4$\\$20\%$&$40.5$&$39.7$&$39.1$&$33.7$&$46.6$&$45.2$&$44.5$&$37.0$&$45.7$&$44.4$&$43.7$&$36.4$\\$30\%$&$42.0$&$41.3$&$40.7$&$35.1$&$47.2$&$45.9$&$45.2$&$37.6$&$46.4$&$45.3$&$44.5$&$37.0$\\$50\%$&$42.7$&$42.2$&$41.5$&$36.3$&$48.3$&$47.2$&$46.4$&$39.0$&$47.3$&$46.4$&$45.6$&$38.3$\\$70\%$&$44.3$&$43.9$&$43.2$&$37.8$&$49.5$&$48.5$&$47.6$&$40.0$&$48.6$&$47.7$&$46.8$&$39.4$\\$90\%$&$43.8$&$43.7$&$42.8$&$37.5$&$47.9$&$47.5$&$46.4$&$39.2$&$46.9$&$46.6$&$45.5$&$38.5$\\$100\%$&$44.3$&$43.9$&$43.1$&$37.7$&$48.4$&$47.5$&$46.6$&$39.3$&$47.4$&$46.6$&$45.8$&$38.7$\\
    \midrule
    \multicolumn{13}{c}{\textbf{EEQA}}\\
    \midrule
    $1\%$&$14.0$&$13.2$&$12.9$&$9.9$&$16.6$&$16.3$&$15.7$&$12.2$&$15.7$&$15.4$&$14.8$&$11.5$\\$3\%$&$20.5$&$18.8$&$18.7$&$14.4$&$23.3$&$22.1$&$21.7$&$17.0$&$22.4$&$21.3$&$20.8$&$16.4$\\$5\%$&$24.0$&$21.5$&$21.7$&$17.0$&$27.1$&$25.0$&$25.0$&$20.0$&$26.1$&$24.2$&$24.0$&$19.2$\\$7\%$&$24.2$&$21.8$&$21.9$&$17.1$&$27.1$&$25.2$&$25.0$&$20.0$&$26.4$&$24.5$&$24.4$&$19.5$\\$10\%$&$25.9$&$23.4$&$23.5$&$18.5$&$29.1$&$27.0$&$26.9$&$21.7$&$28.3$&$26.4$&$26.2$&$21.1$\\$20\%$&$26.7$&$24.5$&$24.5$&$19.6$&$29.9$&$28.2$&$28.0$&$22.9$&$29.1$&$27.4$&$27.2$&$22.2$\\$30\%$&$26.3$&$24.1$&$24.1$&$19.4$&$29.6$&$27.8$&$27.7$&$22.7$&$28.6$&$27.0$&$26.7$&$21.9$\\$50\%$&$27.8$&$25.5$&$25.6$&$20.7$&$31.3$&$29.4$&$29.2$&$24.0$&$30.3$&$28.5$&$28.3$&$23.2$\\$70\%$&$28.1$&$25.7$&$25.8$&$20.9$&$31.4$&$29.5$&$29.3$&$24.2$&$30.5$&$28.7$&$28.5$&$23.5$\\$90\%$&$28.2$&$26.0$&$26.0$&$21.0$&$31.5$&$29.8$&$29.6$&$24.4$&$30.7$&$29.1$&$28.8$&$23.7$\\$100\%$&$28.3$&$26.1$&$26.1$&$21.1$&$31.6$&$29.9$&$29.6$&$24.5$&$30.8$&$29.2$&$28.9$&$23.7$\\
    \midrule
    \multicolumn{13}{c}{\textbf{PAIE}}\\
    \midrule
$1\%$&$33.6$&$32.7$&$31.8$&$25.3$&$39.3$&$38.4$&$37.4$&$30.1$&$39.1$&$38.3$&$37.3$&$30.0$\\$3\%$&$37.0$&$36.0$&$35.2$&$28.6$&$43.2$&$42.0$&$41.1$&$33.7$&$43.1$&$42.0$&$41.1$&$33.7$\\$5\%$&$38.6$&$37.4$&$36.7$&$30.0$&$45.9$&$44.4$&$43.6$&$35.8$&$46.2$&$44.8$&$43.9$&$36.1$\\$7\%$&$39.8$&$39.0$&$38.1$&$31.5$&$45.8$&$45.0$&$43.9$&$36.5$&$46.1$&$45.3$&$44.2$&$36.7$\\$10\%$&$40.6$&$40.0$&$39.0$&$32.4$&$46.9$&$46.2$&$45.1$&$37.6$&$47.2$&$46.6$&$45.4$&$37.9$\\$20\%$&$43.2$&$42.1$&$41.3$&$34.7$&$49.5$&$48.3$&$47.4$&$39.9$&$49.8$&$48.6$&$47.7$&$40.1$\\$30\%$&$43.2$&$42.1$&$41.3$&$34.7$&$49.5$&$48.3$&$47.4$&$39.9$&$49.8$&$48.6$&$47.7$&$40.1$\\$50\%$&$43.4$&$42.6$&$41.8$&$35.3$&$49.9$&$49.0$&$48.0$&$40.6$&$50.3$&$49.4$&$48.4$&$40.9$\\$70\%$&$44.0$&$43.0$&$42.3$&$35.8$&$50.7$&$49.5$&$48.7$&$41.3$&$51.2$&$50.1$&$49.1$&$41.7$\\$90\%$&$44.4$&$43.4$&$42.7$&$36.2$&$51.3$&$49.9$&$49.1$&$41.8$&$51.5$&$50.3$&$49.4$&$42.0$\\$100\%$&$44.5$&$43.4$&$42.7$&$36.3$&$50.8$&$49.4$&$48.7$&$41.4$&$51.1$&$49.8$&$49.0$&$41.7$\\
    
    \bottomrule
    \end{tabular}
    \caption{Experimental results (\%) of the EAE models using different proportions of training data of \ourdata. In this experiment, we adopt the gold trigger evaluation approach~\citep{peng-etal-2023-devil}.}
    \label{tab:app_data_size}
\end{table*}

\subsection{Entity and Non-Entity Arguments}
\label{sec:app_non_entity}
Table~\ref{tab:app_non_entity} presents the overall results on entity and non-entity arguments of \ourdata. The non-entity arguments do not have coreferential relationship with each other and hence there is no entity coreference level evaluation for them.

\begin{table*}[t!]
    \centering
    \small
    \begin{tabular}{l|c|rrrr|rrrr|rrrr}
    \toprule
    \multirow{2}{*}{\textbf{Model}} & \multirow{2}{*}{\textbf{\#Params}}& \multicolumn{4}{c|}{\textbf{Mention Level}} & \multicolumn{4}{c|}{\textbf{Entity Coref Level}} & \multicolumn{4}{c}{\textbf{Event Coref Level}} \\
    & & \multicolumn{1}{c}{\textbf{P}}& \multicolumn{1}{c}{\textbf{R}} & \multicolumn{1}{c}{\textbf{F1}}& \multicolumn{1}{c|}{\textbf{EM}}& \multicolumn{1}{c}{\textbf{P}}& \multicolumn{1}{c}{\textbf{R}} & \multicolumn{1}{c}{\textbf{F1}}& \multicolumn{1}{c|}{\textbf{EM}}& \multicolumn{1}{c}{\textbf{P}}& \multicolumn{1}{c}{\textbf{R}} & \multicolumn{1}{c}{\textbf{F1}}& \multicolumn{1}{c}{\textbf{EM}}\\
    \midrule
    \multicolumn{14}{c}{\textbf{Entity Argument}}\\
    \midrule
    DMBERT & $110$M &$19.7$&$19.7$&$19.7$&$19.5$&$12.5$&$12.4$&$12.4$&$12.3$&$11.8$&$11.8$&$11.8$&$11.6$ \\
    CLEVE & $355$M & $22.1$&$22.1$&$22.1$&$22.0$&$13.2$&$13.2$&$13.2$&$13.0$&$12.3$&$12.2$&$12.2$&$12.1$ \\
    BERT+CRF & $110$M & $18.6$&$18.5$&$18.5$&$17.8$&$12.7$&$12.6$&$12.6$&$12.0$&$12.0$&$11.8$&$11.8$&$11.3$ \\
    EEQA & $110$M & $6.3$&$6.2$&$6.2$&$5.6$&$9.1$&$9.1$&$9.0$&$8.3$&$8.5$&$8.5$&$8.4$&$7.7$ \\
    Text2Event & $770$M & $5.5$&$5.6$&$5.5$&$5.2$&$4.0$&$4.0$&$4.0$&$3.7$&$3.2$&$3.2$&$3.1$&$2.9$ \\
    PAIE & $406$M & $20.4$&$20.5$&$20.3$&$19.2$&$21.0$&$21.1$&$20.9$&$19.8$&$20.0$&$20.1$&$19.9$&$18.9$ \\
    \midrule
    \multicolumn{14}{c}{\textbf{Non-Entity Argument}}\\
    \midrule
    BERT+CRF & $110$M & $24.8$&$24.8$&$24.0$&$19.4$&$-$&$-$&$-$&$-$&$25.3$&$25.3$&$24.5$&$19.6$ \\
    EEQA & $110$M & $18.9$&$17.6$&$17.5$&$13.9$&$-$&$-$&$-$&$-$&$18.6$&$17.4$&$17.2$&$13.7$ \\
    Text2Event & $770$M & $1.7$&$1.7$&$1.6$&$1.1$&$-$&$-$&$-$&$-$&$1.5$&$1.6$&$1.5$&$1.1$ \\
    PAIE & $406$M & $39.4$&$38.4$&$37.6$&$30.4$&$-$&$-$&$-$&$-$&$38.7$&$37.8$&$36.9$&$29.7$ \\
    \bottomrule
    \end{tabular}
    \caption{Experimental results (\%) of existing state-of-the-art fine-tuned EAE models on entity and non-entity arguments of \ourdata. Classification-based models, e.g., DMBERT and CLEVE, are inapplicable to non-entity arguments.}
    \label{tab:app_non_entity}
\end{table*}

\end{document}